# FALSE: Fake News Automatic and Lightweight Solution

Fatema Al Mukhaini, Shaikhah Al Abdoulie, Aisha Al Kharuosi, Amal El Ahmad, Monther Aldwairi
*College of Technological Innovation*
Zayed University
Abu Dhabi, UAE
{201301519, M80008017, M80007973, M80008020, monther.aldwairi}@zu.ac.ae

*Abstract*—Fake news existed ever since there was news, from rumors to printed media then radio and television. Recently, the information age, with its communications and Internet breakthroughs, exacerbated the spread of fake news. Additionally, aside from e-Commerce, the current Internet economy is dependent on advertisements, views and clicks, which prompted many developers to bait the end users to click links or ads. Consequently, the wild spread of fake news through social media networks has impacted real world issues from elections to 5G adoption and the handling of the Covid-19 pandemic. Efforts to detect and thwart fake news has been there since the advent of fake news, from fact checkers to artificial intelligence-based detectors. Solutions are still evolving as more sophisticated techniques are employed by fake news propagators. In this paper, R code have been used to study and visualize a modern fake news dataset. We use clustering, classification, correlation and various plots to analyze and present the data. The experiments show high efficiency of classifiers in telling apart real from fake news.

*Keywords—fake news, click baiting, artificial intelligence machine learning, data visualization*

I. INTRODUCTION

Fake news or misinformation may be defined as news or information that is designed to mimic real information with the sole purpose of deluding users to believe that information and act accordingly. False information is provided online for several reasons to gain political or economic advantage [1].

Fake news existed since the invention of the printing press in the eighteenth century [2]. The recent communications, Internet and social media boom has facilitated the generation and spread of news both real and false. More and more people get their news from social media as opposed to the more trusted traditional news organizations. Social media networks (SMN) motivate and engage people, because they make it easier to cover news, reach consumers faster, spread, search, and finally provide free news [3]. Most print and cable TV media are establishing a significant presence on social media from completely going online to hybrid delivery modes. Most of the investigative journalism, documentaries and real news are very expensive to make and take a long time to prepare and broadcast. On the other hand, any social media user with a smartphone turns into a media reporter in a split second. His news is live, fast spreading, easy to go viral, usually uncensored, real and more credible! Moreover, viewers may instantly comment and interact with the newscaster to learn more about the event being reported [4].

Although, the news quality from conventional news organizations is higher than that of the social media and other online sources, yet the majority of users get their news online and trust it more. It is becoming harder and harder for traditional news organizations that adhere to professional journalism standards to compete with these new light cost and fast spread news organizations that appeal to what the bases want to hear as opposed to reality [5].

On the other side, the Internet economy is based on the sales of targeted advertisements. Tech giants track, store and analyze users' data, clicks and behaviors with the main purpose of delivering customized ads. Even though the advertisements economy keeps evolving and payment schemes may differ, advertising agents are usually compensated on pay-per-click (PPC) or number of views basis [6]. Therefore, they must design a convincing landing page [7] and sometimes bait users to click on ads, some of which may be malicious [8]. Moreover, nowadays you find advertorials, which are articles that look like a legitimate news report with the sole purpose of promoting a product or sometimes lead the readers astray [9]. Even many product reviews are paid today, the same goes for blogs, discussions and comments! Therefore, it is becoming very hard to tell organic information from directed misinformation [10].

The ramifications of the wide spread of fake news on societies and individuals are serious, it may break down the trust and undermine the legitimacy of the real news. We saw that in action, for instance, in the 2016 US elections and during the presidency of Donald Trump, who continued to attack trustworthy news organizations and label them as fake [9]. In addition, we saw state-run media or social bots roll out propaganda or bequeath hoaxes to support a certain political view. Moreover, hoaxes may trick people into forming prejudiced beliefs that are false or unethical [4]. We saw that in the "*alt-right*" and the "*QAnon*" cult like mentality where people are fed falsehood and caught in a vicious feedback loop of misinformation, making false information indistinguishable from truth. This was evident during the current Covid-19 pandemic and vaccines misinformation campaigns [11].

Manual fact checking, fact-finding or blue-penciling is hard, expensive, and time consuming. Compared to the speed at which fake news spread, it is a futile process. By the time fact checkers come up with a verdict, the fake news outlet or promoter would have moved to promote a new lie and the damage has been done. On many instances of viral fake news outbreaks, the court of public opinion renders its verdict and even after the truth is revealed, subjects cannot be vindicated, nor the masses are willing to switch their newly adopted beliefs [12]. Consequently, as the information technology revolution has driven the wide spread of fake news, it also may offer the key for a remedy. Recently, there have been massive efforts to develop artificial intelligence driven solutions to detect fake news on SMN and other sources [13].

The contributions of this paper are threefold: 1) summarize and classify the work in the literature 2) *R* code to study, analyze and visualize the LIAR dataset 3) Perform clustering, classification and correlation on the dataset, and compare different classifiers performance. The next section briefly surveys the literature and provides a glimpse of the work being done. Section III outlines the methodology and dataset while Section IV presents the experimental results, visualization and analysis.

II. RELATED WORK

The below is a brief summary of fake news and the solutions that had been proposed by many researchers classified in terms of the methods or platforms used.

*A. Hybrid Systems*

The research proved hybrid systems are more efficient in fake news detection than the baseline model. The hybrid system was composed of a wide array of modern deception calculation methods. In this research, two main strategies are discussed, which are llinguistic and network approaches. Both the linguistic and network approaches aimed at the classification and identification of fake news spreaders. This this research stated that it is known that most liars hide their identity not to be caught. The research acknowledged the difficulty of designing a hybrid fake news detector [14].

Another work proposed a hybrid architecture model for applying BERT and GPT2 for fake news detection. The researchers concluded with the analysis of the topology of the two approaches, which were designed to make the fake news detection tool. They reported modest accuracies of approximately 92% and 90% for both techniques, respectively Furthermore, the paper recognized the lack of proper unified dataset for studying fake news in SMNs [15].

*B. Fact Checking*

This research was based on an analytical study of the news media with the goal of fake news detection within the political news. It provided a case study to investigate the viabilities of automatic political fact-checking. This experiment had shown that at the time of fact checking, stylistic cues could regulate the reliability of the news. Studies regarding different characteristics of political language and media written language were presented, which can also be classified into a 6-point scale [16].

A more recent study of automated fact checking at the sentence level suggested blocking fake articles should take place only after a third-party expert review. They identified the lack of standard corpora as one of the main hindrances to automated fact checking. Therefore, they published a new corpus of social media posts fact checked by third parties [17].

*C. Fake News Detection in SMN*

This research is mainly based on the detection of fake news on social networking sites, especially Facebook. They performed classification on a dataset comprising 15,500 posts on Facebook and 909,236 users. They also evaluated the users and their behavior on Facebook by creating a structure of active Facebook users per hour and classifying them in two categories, which are hoaxes and non-hoaxes. The researchers performed two sets of experiments using logistic regression and harmonic BLC algorithms. The first measured the correctness of the number of the functional algorithms that are present as training set whereas the second estimated the amount of learning information subject to transfer from one set to another [18].

*D. Linguistic Methods*

This research created a predictive model that classified 1,300,000 new posts on Twitter as reliable or suspicious. In this research, news over social media is categorized as propaganda, satire, hoax, clickbait and verified. It investigated various features and used a neural network architecture, which can automatically classify the verified and non-verified news or posts [19].

The study produced a statistical database of tweets, news portals, posts and retweet per post. Additionally, a figure of the communication network between the types of the news is present in this report. The results show that neural network models verified tweet content, social network connections and classified them as verified and suspicious news [19].

*E. Fake News Challenge*

The first stage of the Fake News Challenge (FNC-1) aimed at addressing fake news detection took place in 2017. This paper offered an in-depth critique of the challenge top three performing systems in terms of: dataset, features, models, results & analysis, metrics, and outcomes. They also made the code and a new dataset available to the public [20]

The FNC-1 challenge shortcomings were also analyzed in particular the cases, which can lead to failure of the models. Then the features are evaluated by the Athene model, which showed F1m performing efficiently. The research established that sophisticated machine learning techniques will be able to find the stance on propositional content and not on lexical features [20]. However, we believe that stance detection is not adequate for fake news challenge as the best possible features are not yet able to solve different cases.

*F. The LIAR Dataset*

This research used a novel hybrid and convolutional neural networks, that integrated text and metadata. It used two detection datasets: automatic fake news detection and the LIAR dataset, which is the latest [21]. The paper analyzed the datasets and mechanisms to enable the researchers to understand the false news detection. Finally, the LIAR dataset may be used for instance differentiation, topic modelling, political NLP research, and argument mining [21].

However, the researchers used secondary analysis and provided quantitative evaluation but no qualitative analysis! Additionally, the analysis compared different models using text features only.

*G. Social Bots Spread of Fake News*

Various websites are established to generate fake news, and these produce around 100 articles each week, among which the majority go unnoticed, but a significant portion goes viral. The researchers emphasized that restricting social bots might help mitigate the spread of misinformation online.

It provided methods like use of API and transforming URLs to aid to constrain social bots from spreading fake news [22]. Classifying these malicious websites using URL features could be an added value to fake news detection systems [23].

*H. Satirical Cues*

Satire is a type of deception where cues are incorporated to fabricate deceptiveness. A satirical hoax is normally executed as a jest to attain success. Researchers indicated that people tend to trust mass media sources to make a news that is accurate and fair. However, a survey in UK showed that the most read newspaper is the least trusted! That had driven readers to rely on alternative information sources, which include social media and blogs [24].

The research analyzed every element of satire and its implementation in the detection of fake news. It provided data-driven linguistic observation features to help in fake news detection. Features such as text processing, feature weighing, humor and grammar could be used to detect satire.

*I. Artificial Intelligence-based Methods*

An outburst of work using artificial intelligence (AI) and machine learning (ML) for fake news detection has been reported in the literature and would be hard to summarize.

The researchers in [25] evaluated three separate datasets: Buzzfeed election dataset, Political news dataset and Bur foot and Baldwin dataset to explore and differentiate styling and content between real, satirical and fake news. The research distinguished the features of news in terms of style, complexity and psychological features. The paper elaborated on the difference between the stylistic features and the complexity of fake news, satirical news, and real news. They concluded that titles are important in detecting fake articles as they contain less nouns and stop-words [25].

Point of view analysis and sentiment classification methods showed that the identification of fake news depends on the language's content. Tweets were collected and probabilistic Naïve Bayes (NB) model was used. The NB model was compared to Neutral Network and Support Vector Machine (SVM) classifiers. While NB was faster, less accurate and often required a vast number of records to acquire good outcomes [26].

In contrast, Cybenko and Cybenko did not believe that artificial intelligence or machine learning would be able to categorize and measure what is true news or fake news [27]. They raised the question of how will the programmers avoid the inheritance of biases in the software? So ethical issues will arise in cyber security as computers decide for humans what is true and what is false [27].

III. METHODOLOGY AND DATASET

Before beginning to suggest solutions for fake news detection, one must decide on what dataset to use from the large number of publicly available repositories. The LIAR dataset includes a combination of metadata with text, which provides a significant improvement during analysis [21].

To help solve this problem we provide an *R* code to study the LIAR dataset and look at the most common counts included within the dataset. The counts in the data are divided in five major counts which are: mostly true counts, false counts, half true counts, barely true counts and pants on fire counts.

*A. Dataset*

The LIAR datasets are collected from political debate, TV ads, Facebook posts, tweets, interview, news' releases and others. The LIAR dataset includes 128,000 human labeled short statements from PolitiFact.com's API, and each statement was evaluated by PolitiFact.com editor for its truthfulness. After initial analysis, they found duplicate labels, and merged the full-flop, half-flip, no-flip into false, half-true and true, respectively [21].

In LIAR, there are six labels for truthfulness ratings: pants-fire, false, barely true, half-true, mostly-true, and true. Moreover, the LIAR dataset includes a mix of democrats, republicans and posts from online social medias. Besides, there is rich meta-data for each speaker including current job, home state, party affiliation and credit history [21].

*B. Attributes Discovery*

The LIAR dataset has 14 attributes, where columns 1 through 8 represent the following.

1. Column 1: the ID of the statement.
2. Column 2: the label.
3. Column 3: the statement.
4. Column 4: the subject(s).
5. Column 5: the speaker.
6. Column 6: the speaker's job title.
7. Column 7: the state info.
8. Column 8: the party affiliation.

On the other hand, columns 9-13 represent the total credit history count, including the current statement, as follows.

9. Column 9: barely true counts.
10. Column 10: false counts.
11. Column 11: half true counts.
12. Column 12: mostly true counts.
13. Column 13: pants on fire counts.
14. Column 14: the context (venue / location of the speech or statement).

*C. Data Preparation*

Data preparation is the second phase in analytics lifecycle after the discovery phase. This phase focus is to explore and organize data deemed useful and to get rid of any data that is not useful. The data is copied into Excel and saved as a "*.csv*" file. The researchers must familiarize themselves with the dataset in order to get the best analysis results. The data is cleansed and processed, where only useful and complete records are kept.

Using *R* code, *str(LIAR1)*, where the dataset name is "*LIAR1*", we view the structure and values of the data's attributes. The functions *head(LIAR)* and *tail(LIAR)* are used to view the first and last six lines of the data to make sure that all records have been copied. The *dim(LIAR)* function views the dimensions of the dataset, that is how many rows and columns are there. Finally, the *summary(LIAR)* function, provides a brief summary of each attribute that are used.

### D. Data Analysis Model

The K-means data mining methodology was used to cluster the data to groups. In addition, classification using decision trees was used to visualize the data in clear way, while correlation between variables was used to see how they affect each other. Bar plots model was built to show the labels in the LIAR dataset. Finally, a 3D plot with regression and scatter plot 3D were used to create a 3D histogram. These 3D diagrams help in displaying the most common subjects that are published. This helped in identifying the most common subjects that can lead to have a higher presence of fake news.

### E. Classification

Seven classifiers: Bayes Net (BN), Naïve Bayes, Meta Classification via Regression (MCR), Rules-JRip (RJR),Trees-J48, Trees Random Forest (TRF) and Trees-Random Trees (TRT), were compared in terms of precession, recall, F-measure, ROC and accuracy. The performance of these classification models is presented in the next Section.

## IV. RESULTS AND ANALYSIS

The LIAR dataset analysis using clustering, correlation, plots and classification, shows that most of the news are considered barely true. It was also noticed that there is significant correlation between some counters such as true counts and half true counts. False counts and pants on fire counts have a lower correlation, while especially true counts and pants on fire counts have the smallest correlation. These results could help as determine the news counter published the most and identify the common subject that are published under this counter.

The following subsections present the experimental results and the analysis of the models discussed in Section III.

### A. Clustering

First, delete all the non-numerical values, then use clustering, in *R*, to isolate the information. K-means was used to create a plot to that shows the five variables which are: barely true counts, false counts, half true counts, mostly true counts and pants on fire counts. Clustering helps in selecting the different attributes and presents the best variables that show good clusters as shown by Figure 1.

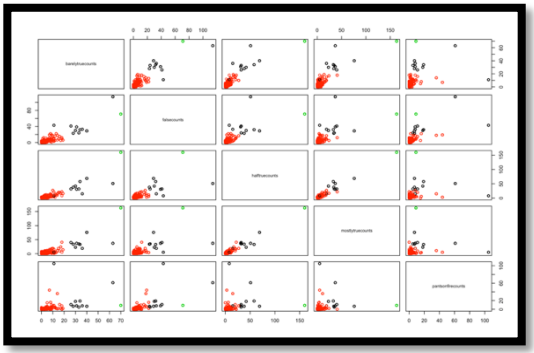

Fig. 1. Clustering plots.

### B. Correlation

Correlation shows the percentage of connection between variables: mostly true counts, false counts, half true counts, barely true counts and pants on fire counts. We have used four different correlation which are.

1. The correlation between barely true counts and half true counts was 0.9136604.
2. The correlation between false counts and pants on fire counts was 0.6610104.
3. The correlation between false counts and mostly true counts was 0.7152603.
4. The correlation between mostly true counts and pants on fire counts was 0.1681535.

This correlation shows that barely true counts and half true counts has a high correlation of 0.9136604, while false counts and pants on fire counts has a lower correlation of 0.6610104. Expectedly, almost no correlation between mostly true counts and pants on fire counts, with a low correlation of 0.1681535.

### C. Bar Plots

Bar plots were used to display the data and make it easier to understand. Command *barplot(table(LIAR1$label))* plots all labels in the dataset as shown in Figure 2.

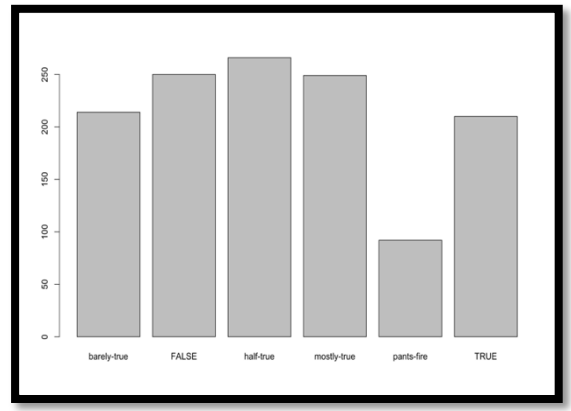

Fig. 2. Labels.

Figure 3 plots top subjects in the database using *barplot(table(LIAR1$subject))* function, while Figure 4, plots the speakers and Figure 5 plots the context. These figures give the researchers a clear idea about the most and least common subjects, speakers and context. Popular subjects expectedly, include flashy topics such as gambling, income, abortion, economy and taxes.

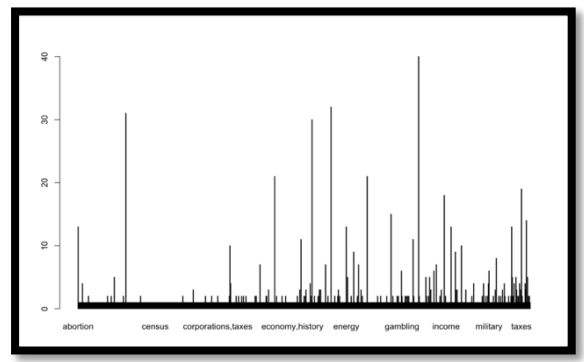

Fig. 3. Subjects.

## D. Classification Model

For a big dataset such as LIAR we use classification. *FSelector* library displays and presents the important attributes in the dataset, *Party* library displays the tree decision of the selected attributes and shows the relation between them. We chose to display the five variables used to count the truthiness of news, which are: mostly true counts, false counts, half true counts, barely true counts and pants on fire counts based on the label. According to the resulting Decision Trees classification, in Figure 6, most labels are considered as barely true.

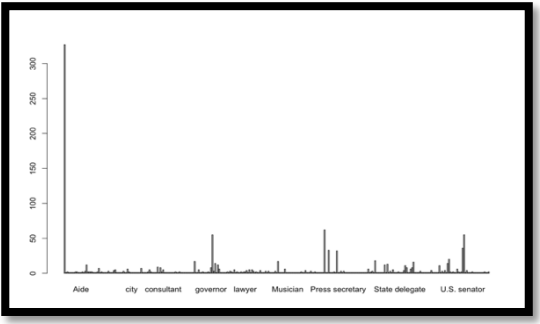

Fig. 4. Speakers.

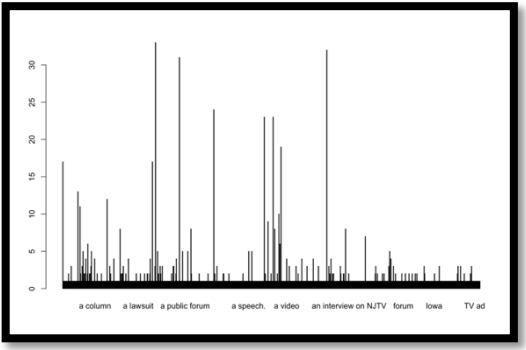

Fig. 5. Context.

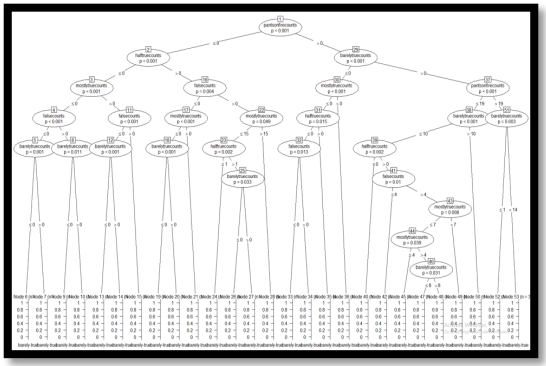

Fig. 6. Decision trees.

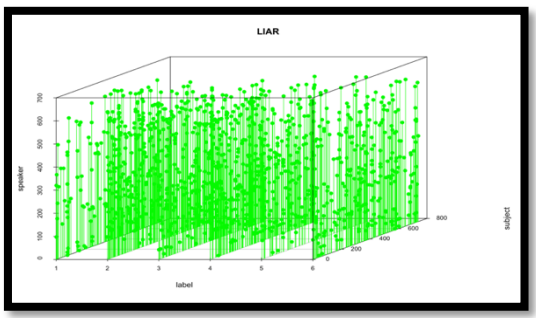

Fig. 7. 3D Model of Label, Subject & speaker.

## E. 3D Scatter Plots

The 3D Scatter plots displays the data in 3D. The 3D scatter plot in Figure 7 shows the three variables: label, subject and speaker.

## F. Classification Results

Table I shows the performance of the difference classifiers in terms of Precision, Recall, F-measure, ROC, and Accuracy. While all classifiers showed excellent results in terms of all metrics, Meta Classification Via Trees, Random Forest and Trees- Random Trees (TRT) performed the best.

TABLE I. CLASSIFICATION RESULTS

| Classifier | Precision | Recall | F-measure | ROC | Accuracy |
|---|---|---|---|---|---|
| BN | 99.8% | 99.4% | 99.6% | 99.9% | 99.4% |
| NB | 99.8% | 99.6% | 98.1% | 99.9% | 96.6% |
| MCR | 99.9% | 99.9% | 99.9% | 99.9% | 99.9% |
| JRip | 99.9% | 99.9% | 99.9% | 99.8% | 99.9% |
| J48 | 99.9% | 99.9% | 99.9% | 99.8% | 99.9% |
| TRF | 99.9% | 99.9% | 99.9% | 99.9% | 99.9% |
| TRT | 99.8% | 99.9% | 99.9% | 99.9% | 99.9% |

## V. CONCLUSIONS

Combating fake news remains a main priority of published media and social media networks alike. We used *R* to study and visualize the LIAR dataset, which is one of the largest most widely used fake news dataset. We performed clustering, classification, as well as correlation to show the connection present between the labels. The data was plotted using bar and 3D plots to make it easier to understood.

The experimental results show high correlation between barely true counts and half true counts and decision trees show a high count of barely fake news category. This presents an even a harder challenge to decern barely true news from those carefully crafted fake news posts.

REFERENCES

[1] M. Aldwairi, and A. Alwahedi, "Detecting Fake News in Social Media Networks," Procedia Computer Science, vol. 141, pp. 215-222, 2018

[2] R. Darnton, The True History of Fake News. The New York Review of Books, 2017

[3] H. Allcott and M. Gentzkow, "Social Media and Fake News in the 2016 Election," The Journal of Economic Perspectives, vol. 31, no. 2, pp. 211-235, 2017.

[4] J. Gorbach, "Not Your Grandpa's Hoax: A Comparative History of Fake News," American Journalism, vol. 35, no. 2, pp. 236-249, 2018.

[5] M. Luo, J.T. Hancock, and D.M. Markowitz. "Credibility Perceptions and Detection Accuracy of Fake News Headlines on Social Media: Effects of Truth-Bias and Endorsement Cues". Communication Research, vol. 49 no. 2, pp. 171-195, 2022

[6] A. Nuara, F. Trovò, N. Gatti, and M. Restelli, "Online Joint Bid/Daily Budget Optimization of Internet Advertising Campaigns," Artificial Intelligence, vol. 305, no. 103663, 2022.

[7] P.K. Keserwani, M.C. Govil, and E.S. Pilli, "The Web Ad-Click Fraud Detection Approach For Supporting to the Online Advertising System," International Journal of Swarm Intelligence, vol. 7, no. 1, pp. 3-24, 2022.

[8] R. Masri and M. Aldwairi, "Automated malicious advertisement detection using VirusTotal, URLVoid, and TrendMicro," In Proceedings of the 8th International Conference on Information and Communication Systems *(ICICS)*, pp. 336-341, Irbid, Jordan, 2017.

[9] S.K. Sharma, S. Kumar, M.S. Manral, and A. Verma, "Paid News Syndrome in Print Media: A Study Based on Selective Newspapers Readers in Jaipur City," Journal of Positive School Psychology, vol. 6 no. 2, 2022.

[10] N. Alsharif, "Fake Opinion Detection in an E-Commerce Business Based on a Long-Short Memory Algorithm," Soft Comput., 2022.

[11] A.R. DiMaggio, "Conspiracy Theories and the Manufacture of Dissent: QAnon, the 'Big Lie', Covid-19, and the Rise of Rightwing Propaganda'," Critical Sociology. 2022.

[12] J.R. Araujo, J.P. Wihbey, and D. Barredo-Ibáñez, "Beyond Fake News and Fact-Checking: A Special Issue to Understand the Political, Social and Technological Consequences of the Battle against Misinformation and Disinformation," Journalism and Media 3, no. 2, pp. 254-256, 2022.

[13] A. Khalil, M. Jarrah, M. Aldwairi, and Y. Jararweh, Detecting Arabic Fake News Using Machine Learning. In Proceedings of the Second International Conference on Intelligent Data Science Technologies and Applications (IDSTA), pp. 171-177, Tartu, Estonia, 15-17 Nov. 2021.

[14] N. Conroy, V. Rubin and Y. Chen, "Automatic Deception Detection: Methods for Finding Fake News," In Proceedings of the 78th ASIS&T Annual Meeting: Information Science with Impact: Research in and for the Community, vol. 1, no. 1, p 85, 2015.

[15] KA. Stevens, S. Chengjie, L. Bingquan and W. Xiaolong, "Transfer Learning and GRU-CRF Augmentation for COVID-19 Fake News Detection,"Computer Science and Information Systems, Mathematics, vol. 10, no. 4, p 585, 2022.

[16] H. Rashkin, E. Choi, J. Jang, S. Volkova and Y. Choi, "Truth of varying shades: Analyzing language in fake news and political fact-checking," In Proceedings of the 2017 Conference on Empirical Methods in Natural Language Processing , vol. 1, no. 1, pp. 2931-2937, 2017.

[17] S. Ahmed, K. Hinkelmann, and F. Corradini, "Fact Checking: An Automatic End to End Fact Checking System". In: Lahby, M., Pathan, AS.K., Maleh, Y., Yafooz, W.M.S. (eds) Combating Fake News with Computational Intelligence Techniques. Studies in Computational Intelligence, vol. 1001. Springer, Cham, 2022.

[18] E. Tacchini, G. Ballarin, M. Della Vedova, S. Moret and L. de Alfaro, "Some like it hoax: Automated fake news detection in social networks," arXiv preprint arXiv:1704.07506, vol. 1, no. 1, 2017.

[19] S. Volkova, K. Shaffer, J. Jang and N. Hodas, "Separating Facts From Fiction: Linguistic Models to Classify Suspicious and Trusted News Posts On Twitter". In Proceedings of the 55th Annual Meeting of the Association for Computational Linguistics, vol. 2, no. 1, pp. 647-653, 2017.

[20] A. Hanselowski, A. PVS, B. Schiller, F. Caspelherr, D. Chaudhuri, C. Meyer and I. Gurevych, "A Retrospective Analysis of the Fake News Challenge Stance Detection Task". In proceedings of the 27th International Conference on Computational Linguistics, Santa Fe, New Mexico, USA, Aug 2018.

[21] W.Y. Wang, ""Liar, Liar Pants on Fire": A New Benchmark Dataset for Fake News Detection". In proceedings of the 55th Annual Meeting of the Association for Computational Linguistics, pp. 422–426, Vancouver, Canada, July 30 - August 4, 2017. 2017.

[22] C. Shao, G.L. Ciampaglia, O. Varol, K.C. Yang, A. Flammini, and F. Menczer, "The spread of low-credibility content by social bots," Nature Communications, vol. 9, no. 4787, 2018.

[23] M. Aldwairi and R. Al-Salman, "MALURLs: Malicious URLs Classification System". In Proceedings of the Annual International Conference on Information Theory and Applications, GSTF Digital Library (GSTF-DL), Singapore, 2011.

[24] V. Rubin, N. Conroy, Y. Chen and S. Cornwell, "Fake News or Truth? Using Satirical Cues to Detect Potentially Misleading News". In Proceedings of the Second Workshop on Computational Approaches to Deception Detection, vol. 1, no. 1, pp. 7-17, 2016.

[25] B. Horne and S. Adali, "This Just in: Fake News Packs a Lot in Title, Uses Simpler, Repetitive Content in Text Body, More Similar to Satire Than Real News". In roceedings of the eleventh International AAAI Conference on Web and Social Media, Montréal, Québec, Canada, May 15–18, 2017.

[26] S. Aphi Wongsophon and P. Chongstitvatana, "Detecting Fake News with Machine Learning Method". In proceedings of the 15th International Conference on Electrical Engineering/Electronics, Computer, Telecommunications and Information Technology (ECTI-CON), pp. 528-531, 2018.

[27] A.K. Cybenko and G. Cybenko, "AI and Fake News," IEEE Intelligent Systems, vol. 33, no. 5, pp. 3-7, 2018.

[28] A. Khalil, M. Jarrah, M. Aldwairi, M. Jaradat, "AFND: Arabic fake news dataset for the detection and classification of articles credibility," Data in Brief, no. 108141, 2022.